\documentclass{article}

\usepackage[preprint]{neurips_2026}
\usepackage[utf8]{inputenc}
\usepackage[T1]{fontenc}
\usepackage{hyperref}
\usepackage{url}
\usepackage{booktabs}
\usepackage{amsfonts}
\usepackage{amsmath}
\usepackage{nicefrac}
\usepackage{microtype}
\usepackage{graphicx}
\usepackage{xcolor}
\usepackage{multirow}
\usepackage{subcaption}
\usepackage{enumitem}

\newcommand{\method}[1]{\textsc{#1}}
\newcommand{\metric}[1]{\texttt{#1}}
\newcommand{\bench}{\textsc{PersonalBench}}

\newcommand{\err}[1]{{\scriptsize$\pm$#1}}

\title{\bench{}: Measuring the Authorship Gap in LLM Personalization}

\author{%
  Yash Ganpat Sawant\\
  Independent AI Researcher\\
  \texttt{sawantyash13@gmail.com}
}

\begin{document}

\maketitle

\begin{abstract}

Personalized text generation aims to make LLMs write in a specific individual's style, yet existing benchmarks measure task accuracy or preference alignment rather than whether the model's output actually resembles the target author's writing. We introduce \bench{}, a benchmark that evaluates inference-time personalization methods through three independent lenses: LUAR (a trained authorship verification model), an LLM-as-judge, and automated stylometrics. Across 50 authors, 1{,}000 generations, and two model families (Qwen~3, GLM-4), we find that personalization methods \emph{do} produce author-differentiated output---LUAR discriminates target authors within generated text at AUC$=$0.918---but this differentiation never crosses the human--LLM boundary. All methods achieve LUAR similarity to real authors in the range 0.484--0.508, below the cross-author human floor of 0.626 (ceiling 0.756). The LLM's own authorship fingerprint dominates: generated text is more distant from any human author than random humans are from each other. Methods are statistically indistinguishable from each other on LUAR (spread 0.024) despite appearing differentiated on the LLM judge---a discrepancy we trace to circularity between trait extraction and profile extraction. We validate that LUAR reliably measures authorship in our corpus (AUC$=$0.76 single-post, 0.96 multi-post). We release \bench{} as a calibrated measuring stick: inference-time personalization modulates the LLM's style but does not bridge the gap to human authorship.\footnote{Code and data: \url{https://github.com/yashsawant22/personalbench}.}

\end{abstract}

\section{Introduction}
\label{sec:intro}

Large language models are increasingly deployed where personalization matters: writing assistants matching a user's voice, customer agents maintaining brand tone, and content tools adapting to individual preferences. These applications share a core requirement: the model must write \emph{like a specific person}. The standard approach is inference-time personalization---conditioning the model via few-shot examples, style profiles, or contrastive prompts at generation time, without modifying model weights.

But does this actually work? Current personalization benchmarks cannot answer this question because they measure the wrong thing. LaMP~\citep{salemi2024lamp} evaluates task accuracy (headline generation, product rating). PersonalLLM~\citep{personalllm2025} measures preference alignment. PRISM~\citep{kirk2024prism} captures stated value preferences. None evaluate whether the generated text \emph{sounds like} the target author---whether the model's underlying authorship fingerprint actually shifts toward the target.

We ask a direct question: \textbf{do inference-time personalization methods change how the model writes, or only what it writes about?} To answer this, we need evaluation that goes beyond surface-level text similarity. We need metrics grounded in authorship science---the decades-old discipline of identifying authors from their writing style.

We introduce \bench{}, a benchmark that evaluates personalization through three independent lenses:

\begin{enumerate}[leftmargin=*, itemsep=1pt]
    \item \textbf{LUAR}~\citep{riverasoto2021luar}: a neural authorship verification model trained on millions of Reddit posts, providing calibrated similarity scores. Our primary metric.
    \item \textbf{LLM-as-judge}: decoupled binary trait matching and same-author assessment, providing interpretable style diagnostics.
    \item \textbf{Automated stylometrics}: function word distributions, punctuation patterns, and lexical overlap.
\end{enumerate}

Our central finding is that \textbf{inference-time personalization produces author-differentiated output that remains trapped in the LLM's own style space}. Within generated text, LUAR discriminates target authors at AUC$=$0.918---personalization \emph{is} doing something. But when comparing generated text to real human text, all four methods score between 0.484 and 0.508 on LUAR---below the cross-author human floor of 0.626 (ceiling 0.756). The LLM's authorship fingerprint is so dominant that personalized output is more distant from the target author than random human authors are from each other. Profile extraction appears to ``win'' on the LLM judge (TMR$=$0.542 vs.\ baseline 0.384), but LUAR shows no corresponding advantage (0.502 vs.\ 0.484), exposing circularity between the judge's trait extraction and the method's profile extraction.

We validate that this is not a measurement failure. LUAR achieves AUC$=$0.76 on single-post author discrimination in our blog corpus, rising to 0.96 with multiple posts. The authorship gap between human and LLM text is real and measurable---inference-time personalization modulates style within the LLM regime but does not bridge it.

Our contributions:

\begin{enumerate}[leftmargin=*, itemsep=2pt]
    \item \textbf{The \bench{} benchmark}: 50 authors, 1{,}000 generations, four methods, three independent evaluation metrics including LUAR authorship verification---the first personalization benchmark grounded in authorship science.
    \item \textbf{A multi-metric evaluation framework}: combining trained authorship embeddings (LUAR), LLM-as-judge assessment, and classical stylometrics, revealing that cross-metric correlations are uniformly near zero ($|r| < 0.07$ across primary metrics; up to $|r| = 0.17$ when including ROUGE-L) and that single-metric evaluation is unreliable.
    \item \textbf{A rigorously validated finding on the human--LLM authorship gap}: inference-time personalization produces author-differentiated output within the LLM's style space (gen$\leftrightarrow$gen AUC$=$0.918) but does not bridge the gap to human authorship (gen$\rightarrow$real AUC$=$0.632), supported by calibrated baselines, two model families, and convergent evidence from three metrics.
\end{enumerate}

\section{Related Work}
\label{sec:related}

\paragraph{Personalization benchmarks.}
LaMP~\citep{salemi2024lamp} evaluates personalized language model predictions across seven tasks using task-specific accuracy. LongLaMP~\citep{kumar2024longlamp} extends this to long-form generation, introducing content-summary prompts to separate style from content---an approach we adopt. PersonalLLM~\citep{personalllm2025} evaluates preference personalization using synthetic diverse user profiles. PRISM~\citep{kirk2024prism} provides real human preference data but targets value alignment. None evaluate authorial style fidelity or use authorship verification models for evaluation.

\paragraph{Authorship verification.}
Computational authorship analysis spans from function-word frequencies to neural methods~\citep{stamatatos2009survey}. LUAR~\citep{riverasoto2021luar} learns universal authorship representations via contrastive learning on Reddit data with strong cross-domain transfer. We adopt LUAR as our primary metric for its calibrated authorship similarity scores.

\paragraph{LLM style evaluation and detection.}
\citet{wang2025catchme} find that LLMs ``still struggle to imitate everyday authors'' across 400+ authors---our LUAR analysis corroborates and quantifies this. Panza~\citep{panza2024} and ExPerT~\citep{expert2025} evaluate personalized generation but target single systems; we provide a benchmark framework with calibrated baselines. The AI-generated text detection literature~\citep{mitchell2023detectgpt, kirchenbauer2023watermark} has established that LLM outputs carry distinctive fingerprints. Our finding of a ``generated-text regime'' below the human floor on LUAR is consistent: the LLM's authorship signal resists erasure by personalization.

\paragraph{LLM-as-judge.}
Prometheus~\citep{kim2024prometheus} found 60--70\% of Likert scores cluster at 3--4; RULERS~\citep{rulers2025} proposed binary evidence-anchored scoring. We build on binary trait matching and reference anchoring~\citep{expert2025}, adding a decoupled protocol that separates trait scoring from holistic authorship judgment.

\section{The \bench{} Benchmark}
\label{sec:benchmark}

\subsection{Data}
\label{sec:data}

We use the Blog Authorship Corpus~\citep{schler2006effects}, containing 681K posts from 19,320 bloggers with demographic metadata. We select 50 authors meeting: (i)~at least 200 training posts, (ii)~at least 50 test posts, and (iii)~average post length $\geq$ 100 words. Posts are split 80/20 per author into training and test sets, yielding 104K training and 26K test posts. We preprocess by stripping URL placeholders (\texttt{urlLink}) and HTML artifacts. The full corpus of 200 qualifying authors is included in the release for scaling experiments; we report results on 50.

\paragraph{Corpus characteristics.} Posts span 1999--2004, covering personal journals, opinion pieces, creative writing, and daily observations. Mean post length is 207 words ($\sigma{=}189$). Authors exhibit substantial style diversity: average sentence length ranges from 8.4 to 24.1 words; type-token ratio from 0.38 to 0.71; function word distributions show distinctive per-author patterns.

\subsection{Prompt Construction and Contamination Ablation}
\label{sec:contamination}

A critical design decision is how to derive writing prompts from test posts without leaking the author's voice. Na\"ive approaches---extracting the first sentence---embed hundreds of characters of the author's raw text into the prompt, including distinctive vocabulary, punctuation, and tonal markers.

We demonstrate this is not a theoretical concern. When prompts are extracted as raw first sentences, the unpersonalized baseline achieves 50\% on same-author judgments. Replacing with LLM-extracted content summaries---neutral descriptions of \emph{what} the post discusses---drops the baseline to 22\%, a \textbf{28 percentage point reduction} (Table~\ref{tab:contamination}). Following \citet{wang2025catchme}, \citet{panza2024}, and \citet{kumar2024longlamp}, we extract content summaries by prompting: ``Summarize this blog post in 1--2 neutral sentences. Describe what it is about, not how it is written. Do not copy the author's phrasing, tone, or vocabulary.''

\begin{table}[t]
\centering
\caption{Effect of prompt construction on baseline scores. Na\"ive extraction leaks author voice, inflating the unpersonalized baseline by 28pp on same-author judgments.}
\label{tab:contamination}
\small
\begin{tabular}{lcccc}
\toprule
\textbf{Prompt type} & \multicolumn{2}{c}{\textbf{Non-personalized}} & \multicolumn{2}{c}{\textbf{Few-shot vs.\ baseline}} \\
\cmidrule(lr){2-3} \cmidrule(lr){4-5}
& TMR & SA\% & $\Delta$TMR & Win\% \\
\midrule
Raw first sentence & 0.587 & 50\% & $-0.080$ & 23\% \\
Content summary & 0.384 & 22\% & $+0.049$ & 33\% \\
\bottomrule
\end{tabular}
\end{table}

\subsection{Personalization Methods}
\label{sec:methods}

We evaluate four inference-time methods spanning a progression from no personalization to explicit quantitative style signals. All methods receive the same content-summary prompt; they differ only in how author information is represented.

\paragraph{\method{Non-Personalized} (control).} The model receives only the content summary with no author information.

\paragraph{\method{Few-Shot} (explicit style transfer).} Five randomly sampled training posts from the target author are presented alongside an explicit system instruction to match the author's writing style, sentence structure, tone, vocabulary, and rhetorical patterns.

\paragraph{\method{Profile Extraction} (explicit abstract style transfer).} A two-stage approach: (i)~the LLM reads up to 10 training posts and produces a structured style profile covering tone, formality, vocabulary, sentence structure, and rhetorical devices; (ii)~the generation model receives \emph{only this abstract profile}---no raw samples---and generates. The profile is extracted once per author and reused across prompts.

\paragraph{\method{Contrastive with Features} (explicit quantitative style transfer).} The model receives the author's samples alongside contrastive examples from three other authors labeled ``avoid these styles,'' plus computed stylometric features: average sentence length, vocabulary richness, and top function word frequencies for both target and contrastive authors.

\subsection{Evaluation Framework}
\label{sec:eval_framework}

We evaluate generated text through three independent lenses, ordered by the strength of their grounding in authorship science.

\subsubsection{LUAR Authorship Similarity (Primary)}
\label{sec:luar}

LUAR (Learning Universal Authorship Representations)~\citep{riverasoto2021luar} is a transformer trained via contrastive learning on millions of Reddit posts to produce author-discriminative text embeddings. Given two texts, we compute cosine similarity between their LUAR embeddings. Higher similarity indicates the texts are more likely to share an author.

We use LUAR as our primary metric because it integrates authorship signal across syntax, lexical choice, and discourse patterns into a single calibrated score validated on authorship verification benchmarks.

\subsubsection{LLM-as-Judge: Decoupled Binary Evaluation (Secondary)}
\label{sec:judge}

Our judge uses three decoupled stages: (1)~\emph{Trait extraction} (cached per author): the judge reads 10 training posts and extracts 5 distinctive style traits as yes/no questions, constrained to writing style only. (2a)~\emph{Trait scoring}: a separate call scores each trait Y/N with evidence, seeing only the generated text. (2b)~\emph{Same-author judgment}: another separate call compares the generated text to the author's reference post, asking ``could these have been written by the same person?'' Stages 2a and 2b are decoupled because we empirically found that combining them causes the holistic question to shift trait answers. We use Qwen~3 32B~\citep{qwen2025qwen3} for generation and GLM-4 32B~\citep{glm2024chatglm} for judging to mitigate self-enhancement bias~\citep{zheng2023judging}.

\paragraph{Derived metrics.}
\begin{itemize}[leftmargin=*, itemsep=1pt]
    \item \metric{Trait Match Rate} (TMR) $= \text{traits\_present} / 5$: interpretable per-trait diagnostic.
    \item \metric{Same-Author Rate} (SA\%): percentage receiving a ``yes'' judgment.
\end{itemize}

\subsubsection{Automated Stylometrics (Tertiary)}

We compute stylometric similarity between generated text and the author's training posts:
\begin{itemize}[leftmargin=*, itemsep=1pt]
    \item \textbf{Function word cosine} (\metric{FuncCos}): cosine similarity over frequency distributions of 60 common function words---established markers of individual writing style~\citep{argamon2003style}.
    \item \textbf{Punctuation cosine}: similarity over distributions of 10 punctuation marks.
    \item \textbf{ROUGE-L}~\citep{lin2004rouge}: longest common subsequence overlap with the reference post.
\end{itemize}

\section{Experiments}
\label{sec:experiments}

\subsection{Setup}
\label{sec:setup}

We evaluate 50 authors $\times$ 5 prompts $\times$ 4 methods $=$ 1{,}000 generations. The generator is Qwen~3 32B 4-bit, served locally via mlx-lm. The LLM judge is GLM-4 32B 4-bit (different model family). LUAR embeddings are computed using the pretrained model from~\citet{riverasoto2021luar}. Total compute: $\sim$3{,}300 LLM calls over $\sim$24 hours on a single Apple M4~Pro (48GB).

\subsection{LUAR Validation: The Task Is Measurable}
\label{sec:luar_validation}

We verify that LUAR can discriminate authorship in our blog corpus despite being trained on Reddit. On 2{,}500 same-author and 2{,}500 cross-author pairs, single-post LUAR achieves \textbf{AUC$=$0.76} (vs.\ TF-IDF baseline AUC$=$0.54); with 5-post aggregation, AUC rises to \textbf{0.96}. Same-author similarity averages \textbf{0.756} (ceiling); cross-author averages \textbf{0.626} (floor). We report 5-post aggregated scores throughout.

\subsection{Main Result: The Human--LLM Authorship Gap}
\label{sec:main_results}

\begin{table}[t]
\centering
\caption{Method comparison across 50 authors, 1{,}000 generations. LUAR similarity uses 5-post aggregation (5 generations per author-method pair compared to 5 training posts) and is the primary authorship metric ($\uparrow$ better). TMR = trait match rate from LLM judge. SA\% = same-author rate. FuncCos = function word cosine similarity. Ceiling = real author's test posts; floor = cross-author random pairs. All methods score \emph{below} the real-text cross-author floor on LUAR despite appearing differentiated on TMR. CIs use hierarchical bootstrap (resampling authors, then generations within authors) to account for within-author correlation ($B{=}10{,}000$).}
\label{tab:main}
\small
\begin{tabular}{lcccc}
\toprule
\textbf{Method} & \textbf{LUAR} $\uparrow$ & \textbf{TMR} $\uparrow$ & \textbf{SA\%} $\uparrow$ & \textbf{FuncCos} $\uparrow$ \\
\midrule
\method{Non-Personalized} & 0.484\err{.019} & 0.384\err{.058} & 22\%\err{7} & 0.741\err{.011} \\
\method{Few-Shot}         & \textbf{0.508}\err{.020} & 0.433\err{.061} & 31\%\err{8} & 0.749\err{.011} \\
\method{Profile Extraction} & 0.502\err{.019} & \textbf{0.542}\err{.060} & 29\%\err{8} & \textbf{0.761}\err{.010} \\
\method{Contrastive}      & 0.494\err{.020} & 0.447\err{.059} & \textbf{36\%}\err{8} & 0.752\err{.011} \\
\midrule
Real Author (ceiling)     & 0.756 & 0.427 & 30\% & --- \\
Cross-Author (floor)      & 0.626 & 0.390 & 7\%  & --- \\
\bottomrule
\end{tabular}
\end{table}

\begin{figure}[t]
\centering
\includegraphics[width=0.75\textwidth]{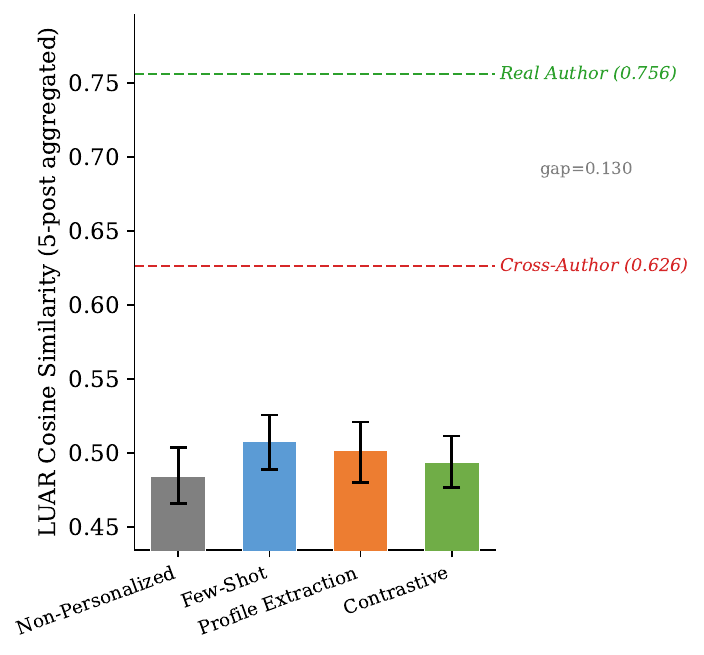}
\caption{LUAR authorship similarity by method (5-post aggregation). All methods score \emph{below} the real-text cross-author floor (0.626), far from the real-author ceiling (0.756). The total spread across methods is only 0.024. The LLM's authorship fingerprint is so dominant that generated text sits in a distinct regime below human-to-human similarity.}
\label{fig:luar_main}
\end{figure}

Table~\ref{tab:main} presents our central finding. On LUAR, all methods score 0.484--0.508 (spread 0.024), \emph{below} the cross-author floor of 0.626. Personalization is not doing nothing---gen$\rightarrow$target (0.497) exceeds gen$\rightarrow$wrong (0.459), AUC$=$0.632---but output remains in the LLM's style space. The LLM judge tells a different story: \method{Profile Extraction} achieves TMR$=$0.542 vs.\ baseline 0.384. We analyze this discrepancy in \S\ref{sec:circularity}. Figure~\ref{fig:luar_main} visualizes the LUAR results.

\subsection{Calibration Analysis}
\label{sec:calibration}

Table~\ref{tab:main} includes calibrated baselines. The \textbf{floor} (cross-author real text) yields LUAR$=$0.626, TMR$=$0.390, SA$=$7\%. The \textbf{ceiling} (same-author real text) yields LUAR$=$0.756, TMR$=$0.427, SA$=$30\%. Notably, the real author scores \emph{lower} on TMR (0.427) than \method{Profile Extraction} (0.542)---evidence of circularity (\S\ref{sec:circularity}). All methods score \emph{below} the real-text floor on LUAR: the best (\method{Few-Shot}, 0.508) is below 0.626. Generated text is more distant from any human author than random humans are from each other. Comparing generated text to wrong vs.\ target authors yields a faint signal (gen$\rightarrow$target 0.497 vs.\ gen$\rightarrow$wrong 0.459, AUC$=$0.632).

\begin{figure}[t]
\centering
\includegraphics[width=0.75\textwidth]{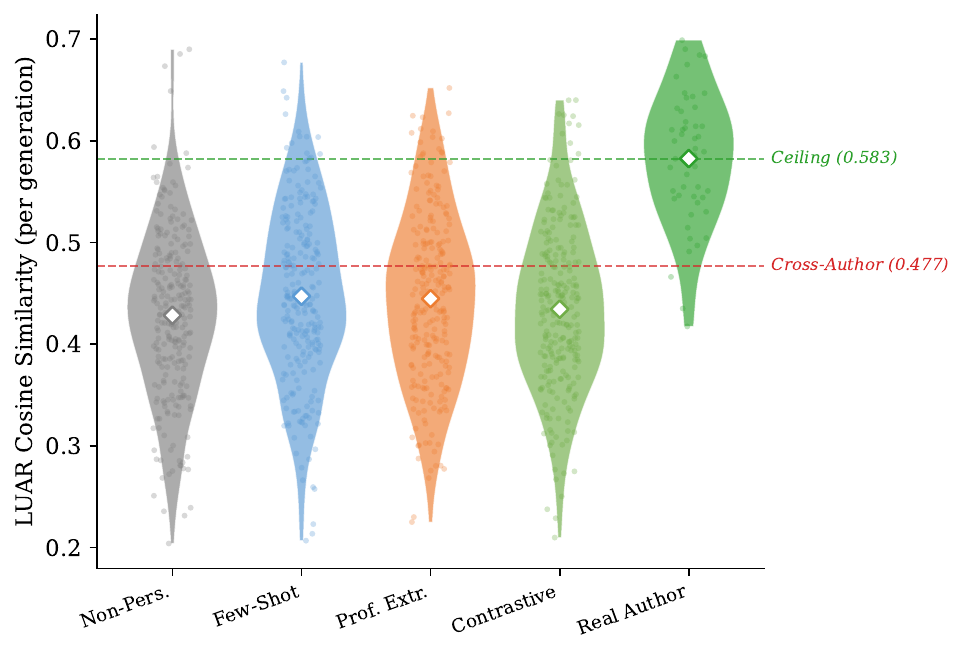}
\caption{Distribution of per-generation LUAR similarity scores by method. All four methods produce overlapping distributions centered near 0.44 (1v5 per-generation scores), well below the real-author distribution centered at 0.58.}
\label{fig:luar_dist}
\end{figure}

\begin{figure}[t]
\centering
\includegraphics[width=\textwidth]{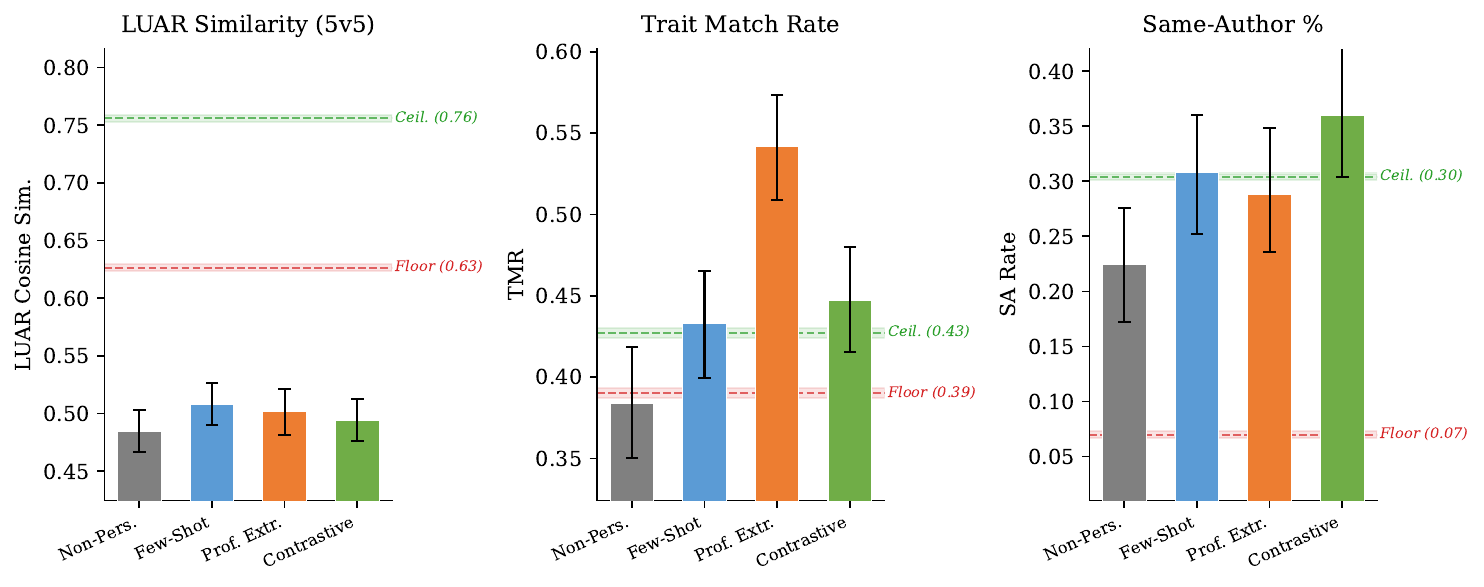}
\caption{Calibration across three metrics. Each panel shows method scores (bars) with ceiling (real author, green) and floor (cross-author, red) baselines. On LUAR (left), all methods score \emph{below} the real-text cross-author floor. On TMR (center), profile extraction exceeds the real-author ceiling---evidence of circularity (\S\ref{sec:circularity}). On SA\% (right), contrastive leads but all methods exceed the chance floor.}
\label{fig:calibration}
\end{figure}

\paragraph{The generated-text regime.} Gen$\leftrightarrow$gen LUAR similarity averages 0.932 (same target) and 0.858 (different targets)---far above gen$\leftrightarrow$real (0.522). Within this cluster, LUAR discriminates target authors at AUC$=$0.918: personalization modulates output within the LLM's style space without crossing into human territory. Figure~\ref{fig:luar_dist} confirms all methods produce overlapping distributions; Figure~\ref{fig:calibration} shows calibration across all three metrics.

\subsection{Cross-Metric Agreement}
\label{sec:cross_metric}

\begin{table}[t]
\centering
\caption{Pearson correlation between evaluation metrics ($n{=}1{,}000$). All pairwise correlations are near zero, indicating the metrics capture fundamentally different constructs. No single metric is a reliable proxy for another.}
\label{tab:correlations}
\small
\begin{tabular}{lccc}
\toprule
& \textbf{LUAR} & \textbf{TMR} & \textbf{FuncCos} \\
\midrule
\textbf{LUAR}    & 1.00  & ---   & --- \\
\textbf{TMR}     & 0.013 & 1.00  & --- \\
\textbf{FuncCos} & 0.026 & 0.067 & 1.00 \\
\bottomrule
\end{tabular}
\end{table}

\begin{figure}[t]
\centering
\includegraphics[width=0.75\textwidth]{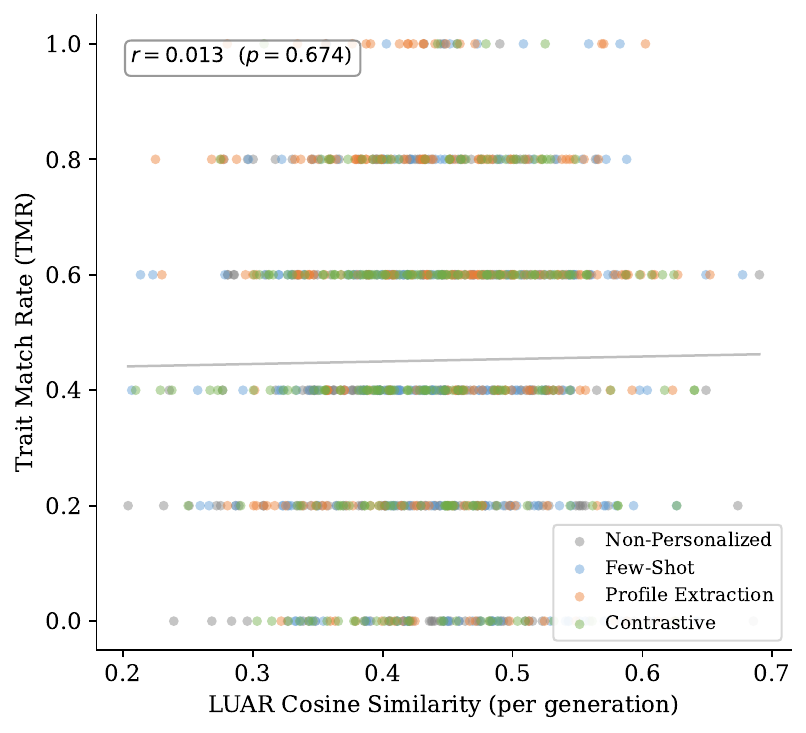}
\caption{LUAR similarity vs.\ TMR for all 1{,}000 generations, colored by method. No systematic relationship ($r{=}0.013$). Profile extraction's high TMR does not correspond to high LUAR.}
\label{fig:cross_metric}
\end{figure}

Table~\ref{tab:correlations} reveals that all pairwise correlations among primary metrics are near zero ($|r| < 0.07$; $|r| \leq 0.17$ including ROUGE-L). A benchmark using only FuncCos would declare \method{Profile Extraction} the winner ($d{=}0.23$); TMR would amplify this ($d{=}0.58$); LUAR would find no differentiation. Single-metric evaluation is unreliable. Figure~\ref{fig:cross_metric} confirms: high TMR does not correspond to high LUAR ($r{=}0.013$), indicating the judge's signal is orthogonal to the trained authorship model.

\subsection{Ablations and Diagnostic Analyses}
\label{sec:ablations}

\subsubsection{TMR Circularity: Why Profile Extraction ``Wins'' on the Judge}
\label{sec:circularity}

\method{Profile Extraction} achieves TMR$=$0.542---far above other methods---but LUAR$=$0.502, indistinguishable from the pack. This discrepancy has a straightforward explanation: \emph{the judge's trait extraction and the method's profile extraction perform the same operation}. Both ask an LLM to read author samples and extract salient style features. The profile is then used to generate text optimized for exactly the kind of features the judge will check.

Supporting evidence: the real author's own text scores TMR$=$0.427---\emph{lower} than \method{Profile Extraction}'s 0.542. If TMR were measuring genuine authorship fidelity, the real author should set the ceiling. Instead, the method that explicitly optimizes for LLM-extractable traits exceeds the real author, indicating TMR measures instruction-following rather than style transfer.

\subsubsection{Trait Extraction Stability}

We extracted traits 3 times per author for a subset of 10 authors and computed pairwise Jaccard similarity between trait sets. Mean Jaccard similarity is \textbf{0.22}---the judge produces substantially different traits on each run. This instability further undermines TMR as a reliable authorship metric: the yardstick itself changes between measurements.

\subsubsection{Additional Ablations}

\paragraph{Classical features.} TF-IDF cosine similarity achieves AUC$=$0.54 on generated text (vs.\ LUAR's 0.76 on real text)---classical features cannot find authorship signal when the LLM's fingerprint dominates.

\paragraph{LUAR aggregation sensitivity.} At every aggregation level (1--5 posts), all methods remain below the real-text floor (e.g., 1-post: 0.345--0.368 vs.\ floor 0.398; 5-post: 0.485--0.508 vs.\ 0.626). The finding is robust to aggregation choice.

\subsubsection{Second Generator}

We replicated with GLM-4 32B (10 authors, 150 generations). All methods score below the human floor (0.417--0.576 vs.\ 0.626, AUC$=$0.671). Method spread is larger (0.16 vs.\ Qwen's 0.024), suggesting gap strength varies by model but its existence does not. Cross-model analysis reveals model-specific fingerprints: within-model similarity is high (Qwen$\leftrightarrow$Qwen 0.918, GLM$\leftrightarrow$GLM 0.839), cross-model lower (0.753), and gen$\rightarrow$real lowest (0.45--0.49).

\section{Discussion}
\label{sec:discussion}

\paragraph{Surface vs.\ deep style.}
Prompting shifts surface features (\method{Profile Extraction} achieves function word cosine 0.761 vs.\ baseline 0.741) but not the authorship fingerprint. Gen$\leftrightarrow$gen analysis shows personalization modulates output within the LLM's style space (AUC 0.814--0.920) without crossing into human territory (gen$\leftrightarrow$real AUC$=$0.664), consistent with \citet{wang2025catchme}. This parallels AI-generated text detection: DetectGPT~\citep{mitchell2023detectgpt} and watermarking~\citep{kirchenbauer2023watermark} exploit distinctive statistical signatures in LLM outputs. The same fingerprint that enables detection resists erasure by inference-time conditioning---the model's authorship signal is architecturally embedded, not contextually malleable.

\paragraph{Why metrics disagree.}
Cross-metric correlations near zero ($|r| < 0.07$) reveal that ``personalization'' is not a single construct. TMR measures instruction-following (explaining \method{Profile Extraction}'s advantage---it optimizes for what TMR measures). FuncCos captures surface lexical alignment. LUAR integrates deep authorship signal across syntax, discourse, and lexical choice. No method moves all three, meaning the methods produce metric-specific artifacts rather than genuine style transfer. Single-metric evaluation of personalization is unreliable.

\paragraph{Instruction strength is not the bottleneck.}
\method{Few-Shot} provides real author examples \emph{and} explicit instructions to match the author's style---the strongest inference-time signal short of weight modification. Yet it achieves LUAR$=$0.508, indistinguishable from \method{Profile Extraction} (0.502) and \method{Contrastive} (0.494). The bottleneck is not instruction quality but the model's capacity to shift its deep generative distribution at inference time.

\paragraph{Implications.}
Closing the authorship gap likely requires training-time adaptation: fine-tuning via LoRA, reinforcement learning with style rewards, or continued pretraining on author corpora. \bench{} provides the measuring stick---floor (0.626) and ceiling (0.756) define the target range. Beyond personalization, our findings suggest LLM outputs carry an indelible authorship fingerprint that prompt engineering cannot mask, with implications for AI text detection and the fundamental limits of prompt-based style control. The gen$\leftrightarrow$gen AUC of 0.918 confirms personalization does leave a trace within the model's output space, but all outputs remain in the LLM's stylistic regime rather than crossing into human territory.

\section{Limitations}
\label{sec:limitations}

\paragraph{Two generators, limited diversity.} We validate the main finding on two model families (Qwen~3 32B and GLM-4 32B), but both are 32B-scale 4-bit quantized models. More diverse generators---different parameter scales, architectures, and pretraining data---remain to be tested.

\paragraph{Single domain.} \bench{} evaluates blog-style writing from the early 2000s. Generalization to modern social media, email, or academic writing is untested.

\paragraph{LUAR domain gap and distributional confound.} LUAR was trained on Reddit, not blogs, and never saw LLM-generated text. Real author text is organic blog writing while generated text responds to content-summary instructions---a format difference that may inflate the gen$\rightarrow$real gap. However, the high gen$\leftrightarrow$gen AUC (0.918)---where all texts share the same format---confirms LUAR detects genuine author signal, and gen$\rightarrow$real AUC (0.632) is above chance.

\paragraph{No human validation and other scope.} Our LLM judge has not been validated against human authorship judgments; trait stability is low (Jaccard$=$0.22) and human agreement studies are needed. All data is English; authorship patterns may differ across languages. Both generator and judge use 4-bit quantization, which may affect subtle stylistic capabilities.

\section{Conclusion}
\label{sec:conclusion}

We introduced \bench{}, the first personalization benchmark grounded in authorship verification. Across 50 authors, 1{,}000 generations, and four methods, LUAR similarity clusters at 0.484--0.508---below the real-text cross-author floor of 0.626. The LLM's fingerprint dominates, yet personalization does leave a trace within the generated-text regime (gen$\leftrightarrow$gen AUC$=$0.918). We release \bench{} as a calibrated measuring stick---floor (0.626) and ceiling (0.756) define the authorship gap---designed for extensibility so researchers can evaluate new methods against these baselines. Closing the gap likely requires training-time approaches; \bench{} provides the target.


\bibliographystyle{plainnat}


\appendix

\section*{NeurIPS Paper Checklist}

\begin{enumerate}

\item {\bf Claims}
    \begin{itemize}
        \item[] Question: Do the main claims made in the abstract and introduction accurately reflect the paper's contributions and scope?
        \item[] Answer: \answerYes{}
        \item[] Justification: The abstract and introduction state three specific contributions (benchmark, multi-metric framework, negative finding), all supported by experiments in \S4.
    \end{itemize}

\item {\bf Limitations}
    \begin{itemize}
        \item[] Question: Does the paper discuss the limitations of the work performed by the authors?
        \item[] Answer: \answerYes{}
        \item[] Justification: \S6 discusses five limitations including two generators with limited diversity, single domain, LUAR domain gap, no human validation, and language/quantization scope.
    \end{itemize}

\item {\bf Theory assumptions and proofs}
    \begin{itemize}
        \item[] Question: For each theoretical result, does the paper provide the full set of assumptions and a complete (correct) proof?
        \item[] Answer: \answerNA{}
        \item[] Justification: This is an empirical benchmark paper with no theoretical results.
    \end{itemize}

\item {\bf Experiments reproducibility}
    \begin{itemize}
        \item[] Question: Does the paper fully disclose all the information needed to reproduce the main experimental results of the paper to the extent that it affects the main claims and/or conclusions of the paper?
        \item[] Answer: \answerYes{}
        \item[] Justification: \S4.1 specifies all models, quantization levels, hardware, and generation counts. All prompts are described in \S3. Code and data will be released.
    \end{itemize}

\item {\bf Open access to data and code}
    \begin{itemize}
        \item[] Question: Does the paper provide open access to the data and code, with sufficient instructions to faithfully reproduce the main experimental results?
        \item[] Answer: \answerYes{}
        \item[] Justification: We release all code (evaluation pipeline, generation scripts, analysis tools, and figure reproduction), generated outputs, and processed data via a public repository (\url{https://github.com/yashsawant22/personalbench}). The Blog Authorship Corpus is publicly available. LUAR is publicly available. The processed 200-author corpus will be hosted on a persistent platform for long-term access.
    \end{itemize}

\item {\bf Experimental setting/details}
    \begin{itemize}
        \item[] Question: Does the paper specify all the training and test details necessary to understand the results?
        \item[] Answer: \answerYes{}
        \item[] Justification: \S3 describes data selection criteria, train/test splits, prompt construction, and method implementations. \S4.1 specifies all model and hardware details.
    \end{itemize}

\item {\bf Experiment statistical significance}
    \begin{itemize}
        \item[] Question: Does the paper report error bars suitably and correctly defined or other appropriate information about the statistical significance of the experiments?
        \item[] Answer: \answerYes{}
        \item[] Justification: Table~2 reports hierarchical bootstrap 95\% CIs ($B{=}10{,}000$) for all metrics. We report calibration baselines (floor and ceiling), LUAR AUC validation, and cross-metric correlations with sample sizes.
    \end{itemize}

\item {\bf Experiments compute resources}
    \begin{itemize}
        \item[] Question: For each experiment, does the paper provide sufficient information on the computer resources needed to reproduce the experiments?
        \item[] Answer: \answerYes{}
        \item[] Justification: \S4.1 reports hardware (Apple M4 Pro, 48GB), total LLM calls ($\sim$3,300), and wall-clock time ($\sim$24 hours).
    \end{itemize}

\item {\bf Code of ethics}
    \begin{itemize}
        \item[] Question: Does the research conducted in the paper conform, in every respect, with the NeurIPS Code of Ethics?
        \item[] Answer: \answerYes{}
        \item[] Justification: The Blog Authorship Corpus is a publicly available research dataset. We do not generate harmful content. Our work evaluates rather than enables impersonation.
    \end{itemize}

\item {\bf Broader impacts}
    \begin{itemize}
        \item[] Question: Does the paper discuss both potential positive societal impacts and negative societal impacts of the work?
        \item[] Answer: \answerYes{}
        \item[] Justification: Our benchmark could theoretically aid impersonation research, but our central negative finding---that inference-time methods fail at authorship transfer---demonstrates that current techniques cannot produce text indistinguishable from a target author. This finding itself reduces perceived impersonation risk. Positive impacts include better evaluation methodology for personalization research and a calibrated framework for measuring future progress.
    \end{itemize}

\item {\bf Safeguards}
    \begin{itemize}
        \item[] Question: Does the paper describe safeguards that have been put in place for responsible use of data and methods?
        \item[] Answer: \answerYes{}
        \item[] Justification: We use a publicly available, anonymized blog corpus. Generated outputs are evaluated, not deployed. The benchmark measures style fidelity rather than providing tools for impersonation.
    \end{itemize}

\item {\bf Licenses for existing assets}
    \begin{itemize}
        \item[] Question: Are the creators of assets used in the paper properly credited and are the license and terms of use explicitly mentioned and properly respected?
        \item[] Answer: \answerYes{}
        \item[] Justification: The Blog Authorship Corpus~\citep{schler2006effects} and LUAR~\citep{riverasoto2021luar} are properly cited. Both are publicly available for research use.
    \end{itemize}

\item {\bf New assets}
    \begin{itemize}
        \item[] Question: Are new assets introduced in the paper well documented and is the responsible use of them discussed?
        \item[] Answer: \answerYes{}
        \item[] Justification: We release the benchmark code, evaluation framework, and generated outputs with documentation and a research-use license.
    \end{itemize}

\item {\bf Crowdsourcing and research with human subjects}
    \begin{itemize}
        \item[] Question: For crowdsourcing experiments and research with human subjects, does the paper include the full text of instructions given to participants and screenshots?
        \item[] Answer: \answerNA{}
        \item[] Justification: No crowdsourcing or human subject research was conducted.
    \end{itemize}

\item {\bf Institutional review board (IRB) approvals or equivalent for research with human subjects}
    \begin{itemize}
        \item[] Question: Does the paper describe potential risks incurred by study participants, whether compensation was adequate given the risks, whether informed consent was obtained, etc.?
        \item[] Answer: \answerNA{}
        \item[] Justification: No human subjects research was conducted.
    \end{itemize}

\item {\bf Usage of LLMs}
    \begin{itemize}
        \item[] Question: Does the paper describe the usage of LLMs if it is an important, original, or non-standard component of the core methods in this research?
        \item[] Answer: \answerYes{}
        \item[] Justification: LLMs are central to this work: Qwen~3 32B is used for text generation across all four personalization methods (\S3.3), and GLM-4 32B serves as the LLM judge (\S3.4.2). Both models, their quantization levels, and serving configurations are fully specified in \S4.1. All prompts are reproduced verbatim in Appendix~A.
    \end{itemize}

\end{enumerate}


\section{Prompt Templates}
\label{app:prompts}

All prompts used in \bench{} are reproduced below verbatim. Variables in curly braces (e.g., \texttt{\{prompt\}}) are filled at runtime.

\subsection{Content Summary Extraction}
\label{app:prompt_summary}

Used to extract a neutral content summary from each test post, following the approach of \citet{wang2025catchme} and \citet{kumar2024longlamp}. The summary specifies \emph{what} the post is about without leaking \emph{how} the author writes.

\begin{quote}\small\ttfamily
System: You are a helpful assistant that summarizes blog post topics in plain, neutral language.\\[4pt]
User: Blog post excerpt:\\
\{snippet\}\\[4pt]
Task: In 1-2 plain sentences, what is this blog post about? State the topic only. Do not analyze the writing style. Do not copy the author's words.
\end{quote}

The summary is then wrapped into a generation prompt: \texttt{Write a short blog post (2-3 paragraphs) about the following: \{summary\}}.

\subsection{Non-Personalized (Control)}
\label{app:prompt_nonpersonalized}

The model receives only the content-summary prompt with no author context:

\begin{quote}\small\ttfamily
Write a short blog post (2-3 paragraphs) about the following: \{summary\}
\end{quote}

\subsection{Few-Shot}
\label{app:prompt_fewshot}

Five randomly sampled training posts are presented with an explicit system instruction to match the author's style.

\begin{quote}\small\ttfamily
System: You are writing in the style of the author shown in the examples below. Match their sentence structure, tone, vocabulary, rhetorical patterns, punctuation habits, and overall voice as closely as possible. A reader who knows this author should find it hard to tell your text was not written by them.\\[4pt]
User: Examples from the author:\\[4pt]
[Example 1]\\
\{sample\_1\}\\[4pt]
[Example 2]\\
\{sample\_2\}\\[4pt]
...\\[4pt]
[Example 5]\\
\{sample\_5\}\\[4pt]
Now write a new post in this author's style on the following topic: \{prompt\}
\end{quote}

\subsection{Profile Extraction (Two-Stage)}
\label{app:prompt_profile}

\paragraph{Stage 1: Profile extraction} (one call per author, cached).

\begin{quote}\small\ttfamily
You are a forensic linguist analyzing writing samples from one author. Produce a detailed style profile that would let someone else replicate their writing without seeing the originals.\\[4pt]
Writing samples:\\
\{samples\_text\}\\[4pt]
Profile must cover:\\
1. TONE: Overall attitude (e.g., confessional, detached, enthusiastic, sardonic)\\
2. FORMALITY: Register and how it shifts within a piece\\
3. VOCABULARY: Complexity, jargon, slang, distinctive word choices\\
4. SENTENCE STRUCTURE: Length patterns, fragments, run-ons, punctuation habits\\
5. RHETORICAL MOVES: How they open posts, transition, close. Use of questions, lists, asides\\
6. PERSPECTIVE: Point of view, self-reference patterns, how they address readers\\
7. QUIRKS: Anything distinctive --- ellipses, parentheticals, specific phrases, capitalizations\\
8. EMOTIONAL TEXTURE: How they handle vulnerability, humor, authority\\[4pt]
Be concrete. Quote specific phrases as examples. This profile will be used WITHOUT access to the original samples.
\end{quote}

\paragraph{Stage 2: Generation from profile} (one call per prompt). The model receives only the abstract profile, no raw samples.

\begin{quote}\small\ttfamily
System: You are writing as the person described in the profile below. Follow EVERY detail in the profile --- tone, sentence structure, quirks, vocabulary level, emotional register. The reader should not be able to tell this wasn't written by the original author.\\[4pt]
AUTHOR PROFILE:\\
\{cached\_profile\}\\[4pt]
User: \{prompt\}
\end{quote}

\subsection{Contrastive with Features}
\label{app:prompt_contrastive}

The model receives the target author's samples, contrastive examples from other authors, and computed stylometric features.

\begin{quote}\small\ttfamily
System: You must write in the TARGET author's style, NOT the other authors' styles. Pay close attention to the measurable differences between them.\\[4pt]
QUANTITATIVE STYLE COMPARISON:\\
Target author --- avg sentence length: \{sent\_len\} words, vocabulary richness: \{vocab\}, top function words: \{top\_func\_str\}\\
Other authors --- avg sentence length: \{contrast\_sent\_len\} words, vocabulary richness: \{contrast\_vocab\}\\[4pt]
TARGET AUTHOR writes like this:\\[4pt]
>>> \{author\_sample\_1\}\\
>>> \{author\_sample\_2\}\\
>>> \{author\_sample\_3\}\\[4pt]
OTHER AUTHORS write like this (AVOID these styles):\\[4pt]
>>> \{contrast\_sample\_1\}\\
>>> \{contrast\_sample\_2\}\\
>>> \{contrast\_sample\_3\}\\[4pt]
Key differences to maintain: match the target's sentence length, vocabulary level, and function word patterns. If the target uses shorter sentences than others, keep yours short. If they use more personal pronouns, do the same.\\[4pt]
User: \{prompt\}
\end{quote}

\subsection{Judge Stage 1: Trait Extraction}
\label{app:prompt_judge_traits}

One call per author, cached. Extracts 5 distinctive style traits as yes/no checkable questions.

\begin{quote}\small\ttfamily
You are a forensic linguist. Analyze these writing samples from ONE author and identify exactly 5 distinctive STYLE traits.\\[4pt]
\#\# Writing Samples\\
\{author\_samples\}\\[4pt]
\#\# Instructions\\
For each trait, write it as a YES/NO question that can be checked against new text on ANY topic.\\[4pt]
Focus ONLY on HOW the author writes, not WHAT they write about:\\
- Sentence structure patterns (fragments, run-ons, length)\\
- Opening/closing strategies\\
- Rhetorical devices (questions, lists, asides, parentheticals)\\
- Tone shifts within a piece\\
- Vocabulary level and word choices\\
- Self-reference and reader-address patterns\\
- Emotional register and how they handle vulnerability/humor\\[4pt]
Do NOT include traits about:\\
- URL formatting, link patterns, or platform markup\\
- Generic qualities like "uses complete sentences" or "writes in English"\\
- Specific topics the author writes about\\
- Content preferences --- only writing STYLE\\[4pt]
Return exactly 5 traits. Respond in this exact JSON format:\\
\{"traits": [\{"label": "short trait name", "question": "Does the text ...? (yes/no checkable)"\}]\}
\end{quote}

\subsection{Judge Stage 2a: Trait Scoring}
\label{app:prompt_judge_scoring}

One call per generation. Receives only the generated text and the 5 trait questions---no reference post, no holistic question.

\begin{quote}\small\ttfamily
Check whether generated text exhibits specific style traits.\\[4pt]
\#\# Generated text\\
\{generated\_text\}\\[4pt]
\#\# Style traits to check\\
\{traits\_formatted\}\\[4pt]
\#\# Instructions\\
For EACH trait, answer Y or N. Quote brief evidence from the generated text, or explain why the trait is absent.\\[4pt]
Respond in this exact JSON format:\\
\{"traits": [\{"q": 1, "present": true, "evidence": "brief quote or explanation"\}, \{"q": 2, "present": false, "evidence": "brief explanation"\}]\}
\end{quote}

\subsection{Judge Stage 2b: Same-Author Judgment}
\label{app:prompt_judge_sameauthor}

One call per generation. Receives only the reference post and generated text---no trait questions.

\begin{quote}\small\ttfamily
You are evaluating whether two texts were written by the same person.\\[4pt]
\#\# Text A (reference --- known author)\\
\{reference\_post\}\\[4pt]
\#\# Text B (to evaluate)\\
\{generated\_text\}\\[4pt]
\#\# Instructions\\
Based on writing style alone (not topic), could Text B plausibly have been written by the same person who wrote Text A?\\[4pt]
Consider: sentence structure, vocabulary level, tone, rhetorical devices, emotional register, punctuation habits, how they open and close, self-reference patterns. Ignore topic differences --- focus only on whether the VOICE is the same.\\[4pt]
Answer YES or NO, then give a one-sentence explanation.\\[4pt]
Respond in this exact JSON format:\\
\{"same\_author": true, "reason": "one sentence explanation"\}
\end{quote}

\section{Example Generations}
\label{app:examples}

We present two example generations illustrating the authorship gap. In both cases, methods produce fluent, on-topic text that adopts some surface features of the author's style, but the underlying voice remains distinctively LLM-generated.

\paragraph{Example 1.} Author writes informal, link-heavy retrospective posts with direct reader address.

\noindent\textbf{Content summary:} \emph{A retrospective look at entries from one year ago, highlighting various topics the author wrote about at that time.}

\noindent\textbf{Reference (real author):}
\begin{quote}\small
Journey Inside My Mind - One Year Ago. One of the great things about blogging is that, over time, one develops a history to which he or she can return. With that in mind, let's take a Journey Inside My Mind from one year ago: [links to posts on downloads, compliments, cartoons, Cincinnati Reds, Frank Oz, self-injury, and old friends]
\end{quote}

\noindent\textbf{Profile Extraction:}
\begin{quote}\small
Cheers!!! I figured I'd take some time today to look back at what I wrote exactly one year ago---I mean, time flies, right? Just flipping through the archives reminded me how much growth and change happen in such a short amount of time. [...]
\end{quote}

\noindent\textbf{Non-Personalized:}
\begin{quote}\small
Looking Back: A Year of Reflection. A year ago today marked the start of a journey filled with exploration, learning, and personal growth---all captured through thoughtful blog posts. [...]
\end{quote}

\noindent Profile Extraction captures informal tone (``Cheers!!!'', ``I mean'') but produces generic reflective content. The real author's distinctive list-of-links format and direct ``let's take a Journey'' address are absent from all methods. Non-Personalized defaults to a polished editorial voice that no human blogger would use.

\paragraph{Example 2.} Author writes devotional posts mixing poetry quotation with liturgical observations.

\noindent\textbf{Content summary:} \emph{A reflection on a poem about reverence and redemption, paired with personal and date-related notes.}

\noindent\textbf{Reference (real author):}
\begin{quote}\small
Reverence, by Venerable John Henry Newman, C.O. ``I bow at Jesu's name, for 'tis the Sign / Of awful mercy towards a guilty line. / Of shameful ancestry, in birth defiled [...]'' [followed by personal notes and date]
\end{quote}

\noindent\textbf{Profile Extraction:}
\begin{quote}\small
Today is the First Sunday of Advent---the beginning of a new liturgical year. It strikes me how often this season carries both anticipation and quiet mourning [...] I came across a small, anonymous poem tucked into an old prayer book [...]
\end{quote}

\noindent\textbf{Few-Shot:}
\begin{quote}\small
Reflections on Reverence and Redemption. In the quiet rhythm of daily life, moments often unfold that whisper truths about the divine---truths captured beautifully in poems that blend reverence with redemption. [...]
\end{quote}

\noindent Profile Extraction produces liturgically-themed text (``First Sunday of Advent'') that matches the author's domain but fabricates a poem rather than quoting one. Few-Shot produces generic devotional prose. The real author's distinctive practice of quoting a specific poem with full attribution is absent from all methods.

\section{Additional Materials}
\label{app:additional}

Per-author LUAR breakdowns and full generation outputs are available in the code repository.

\end{document}